\documentclass{article}
\usepackage{spconf,graphicx}
\usepackage{cite}
\usepackage{amsmath,amssymb,amsfonts}
\usepackage{enumerate}
\usepackage{bm}
\usepackage{multirow}
\usepackage{url}

\title{Multi-color balance for color constancy}
\name{Teruaki Akazawa, Yuma Kinoshita and Hitoshi Kiya
}
\address{
Tokyo Metropolitan University \\
Department of Computer Science \\
6--6 Asahigaoka, Hino-shi, Tokyo, 191--0065 Japan
}
\begin{document}
\maketitle
\begin{abstract}
In this paper, we propose a novel multi-color balance adjustment for color constancy.
The proposed method, called ``n-color balancing,'' allows us not only to perfectly correct n target colors on the basis of corresponding ground truth colors but also to correct colors other than the n colors.
In contrast, although white-balancing can perfectly adjust white, colors other than white are not considered in the framework of white-balancing in general.
In an experiment, the proposed multi-color balancing is demonstrated to outperform both conventional white and multi-color balance adjustments including Bradford's model.
\end{abstract}
\section{Introduction}
\label{sec:intro}
Image segmentation and object recognition are required to decompose an image into meaningful regions.
A typical approach to this problem assigns a single class to each pixel in an image.
However, such hard segmentation is far from ideal when the distinction between meaningful regions is ambiguous, such as in the cases of objects with motion blur or color distortion due to illumination and image enhancement\cite{Color_Constancy_effect_on_DNN_2019_ICCV,Fast_Soft_Segmentation_2020_CVPR,kinoshita-san-Hue,kinoshita-san-APSIPA-HUE,kinoshita-san-Semantic_2019}.
Accordingly, we aim to reduce illumination effects on all colors in an image.

A change in illumination affects the pixel values of an image taken with an RGB digital camera because the values are determined by spectral information such as the spectra of illumination.
In the human visual system, it is well known that illumination changes (i.e., lighting effects) are reduced, and this ability keeps the entire color perception of a scene constant.
In contrast, since cameras do not essentially have this ability, white-balancing is applied to images\cite{ChengsBeyondWhite}.

To apply white-balancing to images, we need two steps: estimating a white region with remaining lighting effects (i.e., source white) and mapping the estimated white region into the ground truth white without lighting effects.
Many studies have focused on estimating source white in images\cite{Max_RGB,Grey_World_Theory,Grey_Edge,Cheng_PCA,APAP_bias_illumination_estimation,Deep_WB_Editing}.
However, even when we correctly estimate white regions, colors other than white still include lighting effects.
Therefore, methods for reducing lighting effects for all colors have been investigated as in\cite{vonKriesOriginal,BradfordOriginal,Maloney:color_constacy,Non_diagonal_Color_Correction,Non_diagonal_Color_Correction_PCA_Analysis,ChengsBeyondWhite,CAM16original,Karaimer_2018_CVPR,Akazawa_3CB}.
von Kries's\cite{vonKriesOriginal} and Bradford's\cite{BradfordOriginal} color appearance models were proposed to improve this problem under the framework of white-balancing.
Cheng el at.\cite{ChengsBeyondWhite} also proposed a multi-color balance method. However, the multi-color balancing cannot be applied to von Kries's and Bradford's models even when white is chosen as one of multiple colors, so it does not outperform state-of-the-art white-balancing enough.

In this paper, we propose a novel multi-color balance method, called ``n-color balancing,'' for correcting all colors.
The proposed color balancing is carried out by using n matrices; in comparison, the conventional white and multi-color balance methods are formulated with a single matrix.
In the proposed method, weights among n matrices are also introduced for correcting all colors.

In an experiment, the ColorChecker dataset prepared by Hemrit et al.\cite{ColorCheckerRECommended} is used, and the proposed method is demonstrated to outperform the conventional color balance adjustments.
\section{Related work}\label{sec:related_work}
We summarize conventional methods for color constancy and problems with these methods.
\subsection{Lighting effects on pixel values}
On the basis of the Lambertian model\cite{Lambertian_model}, pixel values of an image taken with an RGB digital camera are determined by using three elements: spectra of illumination $E(\lambda)$, spectral reflectance of objects $S(\lambda)$, and camera spectral sensitivity $R_{C}$ for color $C \in \{{\rm{R}},{\rm{G}},{\rm{B}}\}$, where $\lambda$ spans the visible spectrum in the range of $[380,780]$.
A pixel value $\bm{P}_{\rm{RGB}}=(P_{\rm{R}},P_{\rm{G}},P_{\rm{B}})^\top$ in the camera RGB color space is given by
\begin{equation}
\label{eqn:pixcel_definition}
\bm{P}_{\rm{RGB}} = \int^{780}_{380} \!\!\
      \!{E(\lambda)}
        {S(\lambda)}
        {R_{C}(\lambda)} \:
        {d\lambda}.
\end{equation}
Eq. (\ref{eqn:pixcel_definition}) means that a change in illumination $E(\lambda)$ affects the pixel values in an image.
In the human visual system, the changes (i.e., lighting effects) are excluded, and the overall color perception is constant regardless of illumination differences, known as chromatic adaption.
For mimicking this human ability as a computer vision task, white-balancing is typically performed as a color adjustment technique.
\subsection{White balance adjustment}
By using white-balancing, lighting effects on a white region in an image are accurately corrected if the white region under illumination is correctly estimated.
Many studies on color constancy have focused on correctly estimating white regions\cite{Max_RGB,Grey_World_Theory,Grey_Edge,Cheng_PCA,APAP_bias_illumination_estimation,Deep_WB_Editing}.

White-balancing is performed by
\begin{equation}
\label{eqn:chromadaptWB}
\bm{P}_{\rm{WB}} = \bf{{M}_{\rm{WB}}} \bm{{P}}_{\rm{XYZ}}
,
\end{equation}
where ${\bm{P}_{\rm{XYZ}}=(X_{\rm{P}},Y_{\rm{P}},Z_{\rm{P}})^\top}$ is a pixel value of an image in the XYZ color space, and ${\bm{P}_{\rm{WB}}=(X_{\rm{WB}},Y_{\rm{WB}},Z_{\rm{WB}})^\top}$ is that of a white-balanced image\cite{brucelindbloom}.
$\bf{M}_{\rm{WB}}$ in (\ref{eqn:chromadaptWB}) is given as
\begin{equation} 
\label{eqn:Mwb}
    \bf{M}_{\rm{WB}} = {\bf{M}_{\rm{A}}}^{\rm{-1}}
    \left(
    \begin{array}{cccc}
    {\rho_{\rm{D}}}/{\rho_{\rm{S}}} & 0 & 0 \\
   0 & {\gamma_{\rm{D}}}/{\gamma_{\rm{S}}} & 0 \\
   0 & 0 & {\beta_{\rm{D}}}/{\beta_{\rm{S}}} \\
    \end{array}
    \right)
    \bf{M}_{\rm{A}}
.
\end{equation}
$\bf{M}_{\rm{A}}$ with a size of 3$\times$3 is decided in accordance with an assumed chromatic adaption model\cite{brucelindbloom}.
$(\rho_{\rm{S}},\gamma_{\rm{S}},\beta_{\rm{S}})^\top$ and $(\rho_{\rm{D}},\gamma_{\rm{D}},\beta_{\rm{D}})^\top$ are calculated from a source white point $(X_{\rm{S}},Y_{\rm{S}},Z_{\rm{S}})^\top$ in an input image and a ground truth white point $(X_{\rm{D}},Y_{\rm{D}},Z_{\rm{D}})^\top$ as
\begin{equation}
\label{eqn:color-xfer}
\left(\!
\begin{array}{cccc}
   {\rho_{\rm{S}}} \\
   {\gamma_{\rm{S}}} \\
   {\beta_{\rm{S}}} \\
  \end{array} 
\!  \right)
  = \bf{M}_{\rm{A}} \left(\!
  \begin{array}{cccc}
   {X_{\rm{S}}} \\
   {Y_{\rm{S}}} \\
   {Z_{\rm{S}}} \\
  \end{array}
\! \right) \: {\rm{,and}} \:
 \left(\!
\begin{array}{cccc}
   {\rho_{\rm{D}}} \\
   {\gamma_{\rm{D}}} \\
   {\beta_{\rm{D}}} \\
  \end{array} 
\!  \right)
  = \bf{M}_{\rm{A}} \left(\!
  \begin{array}{cccc}
   {X_{\rm{D}}} \\
   {Y_{\rm{D}}} \\
   {Z_{\rm{D}}} \\
  \end{array}
\!    \right) 
.
\end{equation}
We call using the 3$\times$3-identity matrix as $\bf{M}_{\rm{A}}$ ``white-balancing with XYZ scaling'' in this paper.
Otherwise, von Kries's\cite{vonKriesOriginal} and Bradford's\cite{BradfordOriginal} color appearance models were also proposed for reducing lighting effects on all colors under the framework of white-balancing.
For example, under the use of Bradford's model, $\bf{M}_{\rm{A}}$ is given as
\begin{equation}  
\label{eqn:Ma} 
  \bf{M}_{\rm{A}} = \left(
  \begin{array}{cccc}
   0.8951 & 0.2664 & -0.1614 \\
   -0.7502 & 1.7135 & 0.0367 \\
   0.0389 & -0.0685 & 1.0296 \\
  \end{array}
    \right)
.
\end{equation}
\subsection{Problem with conventional color constancy methods}
White-balancing is a technique that maps a source white point in an image into a ground truth one as in (\ref{eqn:Mwb}).
In other words, colors other than white are not considered in this mapping, although the goal of color constancy is to remove lighting effects on all colors.
To improve this problem, various color appearance models such as those of von Kries\cite{vonKriesOriginal}, Bradford\cite{BradfordOriginal}, and the latest CAM16\cite{CAM16original} have been proposed to reduce lighting effects on all colors under the framework of white-balancing.
In addition, Cheng et al. proposed a method\cite{ChengsBeyondWhite} for considering both achromatic and chromatic colors to further relax the limitation that white-balancing has. 
In their method, multiple colors are used instead of white. However, this method has two problems. 
The first is that it cannot be combined with excellent white-balancing techniques such as Bradford's model, even when white is chosen as one of the multiple colors. 
The other is that selected multiple colors cannot be perfectly adjusted due to the minimization of errors between multiple colors and the ground truth ones.
\section{Proposed method}
To improve problems with conventional methods, a novel multi-color balance method called ``n-color balancing'' is proposed here.
\subsection{Overview of proposed method}
\begin{figure}[tb]
\begin{center}
\includegraphics[keepaspectratio, scale=0.27]{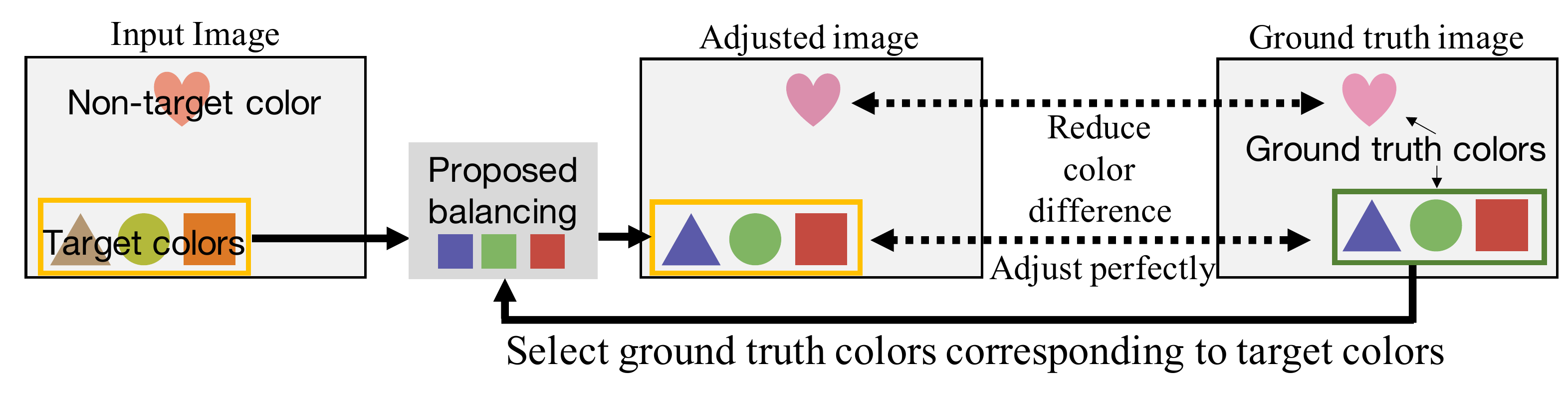}
\end{center}
\vspace{-0.5cm}
\caption{Overview of proposed method (n = 3).}
\label{fig:usage_MCB}
\end{figure}
Figure \ref{fig:usage_MCB} shows an overview of the proposed multi-color balancing. 
We assume that n specific colors, called ``target colors,'' are chosen, and corresponding ground truth colors are given such that there is a color rendition chart in an image. Under the assumption, we aim not only to adjust target colors to the corresponding ground truth colors but also to reduce lighting effects on colors other than the target ones. 
To achieve these objectives, we propose a novel n-color adjustment with n matrices. 
When we choose white as a target color under n = 1, the proposed balancing is reduced to white-balancing.
\subsection{Proposed n-color balancing}
For overall color correction, n-color balancing is carried out by using n matrices, and the matrices are combined with weights; in comparison, the conventional methods including Cheng's multi-color balancing are carried out with a single matrix. 
An n-color balanced pixel ${\bm{P}_{\rm{NCB}}}$ is given by
\begin{equation}
\label{eqn:chromadaptMLT}
\bm{P}_{\rm{NCB}} = {\bf{M}_{\rm{NCB}}} {\bm{P}}_{\rm{XYZ}}
,
\end{equation}
where
\begin{equation}
\label{eqn:M_MLT}
{\bf{{M}_{\rm{NCB}}}} = k_{1} {\bf{M}}_{1} + k_{2} {\bf{M}}_{2} + \cdots + k_{n} {\bf{M}}_{n}.
\end{equation}
$n$ is the number of target colors, and $k_{m}$ is a weight of ${\bf{M}}_{m}$. In (\ref{eqn:M_MLT}), ${\bf{M}}_{m}$, $m \in \{1,2,\cdots,n\}$, is designed in a similar way to white-balancing. 
Let ${\bm{T}}_{m}=(X_{{\rm{T}}m},Y_{{\rm{T}}m},Z_{{\rm{T}}m})^\top$ and ${\bm{G}}_{m}=(X_{{\rm{G}}m},Y_{{\rm{G}}m},Z_{{\rm{G}}m})^\top$ be a target color and a ground truth color in the XYZ color space, respectively. 
${\bf{M}}_{m}$, $m \in \{1,2,\cdots,n\}$, is then given as
\begin{equation} 
\label{eqn:Mm}
    {\bf{M}}_{m} = {\bf{M}_{\rm{A}}}^{\rm{-1}}
    \left(
    \begin{array}{cccc}
    {\rho'_{\rm{D}}}/{\rho'_{\rm{S}}} & 0 & 0 \\
   0 & {\gamma'_{\rm{D}}}/{\gamma'_{\rm{S}}} & 0 \\
   0 & 0 & {\beta'_{\rm{D}}}/{\beta'_{\rm{S}}} \\
    \end{array}
    \right)
    \bf{M}_{\rm{A}}
,
\end{equation}
where
\begin{equation}
\label{eqn:color-xfer_Mm}
\left(\!\!\!
\begin{array}{cccc}
   {\rho'_{\rm{S}}} \\
   {\gamma'_{\rm{S}}} \\
   {\beta'_{\rm{S}}} \\
  \end{array} 
\!\!\!  \right)
 \!\! = \! \bf{M}_{\rm{A}} \left(\!\!\!
  \begin{array}{cccc}
   {X_{{\rm{T}}m}} \\
   {Y_{{\rm{T}}m}} \\
   {Z_{{\rm{T}}m}} \\
  \end{array}
\!\!\!\!  \right) \: {\rm{, and}} \:
 \left(\!\!\!
\begin{array}{cccc}
   {\rho'_{\rm{D}}} \\
   {\gamma'_{\rm{D}}} \\
   {\beta'_{\rm{D}}} \\
  \end{array} 
\!\!\!  \right)
\!\!\!  = \! \bf{M}_{\rm{A}} \left(\!\!\!
  \begin{array}{cccc}
   {X_{{\rm{G}}m}} \\
   {Y_{{\rm{G}}m}} \\
   {Z_{{\rm{G}}m}} \\
  \end{array}
\!\!\!\!  \right) 
.
\end{equation}
Similarly to white-balancing, when choosing white as a target color under $n=1$, n-color balancing is reduced to white-balancing. 
As mentioned above, the use of the 3$\times$3-identity matrix as $\bf{M}_{\rm{A}}$ is n-color balancing with XYZ scaling in this paper. 
Other models such as those by von Kries and Bradford can be also used as $\bf{M}_{\rm{A}}$.

When ${\bm{P}}_{\rm{XYZ}}$ has a color closer to the target color ${\bm{T}}_{m}$ than other target colors, ${\bf{M}}_{m}$ designed with ${\bm{T}}_{m}$ should more contribute to adjusting ${\bm{P}}_{\rm{XYZ}}$ than other matrices. 
Hence, to measure the color similarity between two points, the distance between ${\bm{P}}_{\rm{XYZ}}$ and ${\bm{T}}_{m}$ is calculated from
\begin{equation}
\label{eqn:XYZ_vector_distance}
d_{m} = \sqrt{\left(\frac{X_{\rm{P}}} {Y_{\rm{P}}} - \frac{X_{{\rm{T}}m}} {Y_{{\rm{T}}m}}\right)^{2} + \left(\frac{Z_{\rm{P}}} {Y_{\rm{P}}} - \frac{Z_{{\rm{T}}m}} {Y_{{\rm{T}}m}}\right)^{2}}
.
\end{equation}
A smaller $d_{m}$ means that ${\bm{P}}_{\rm{XYZ}}$ has a closer color to ${\bm{T}}_{m}$. 
Because $k_{m}$ in (\ref{eqn:M_MLT}) should be larger under a smaller $d_{m}$, the inverse proportion to $d_{m}$ is calculated as
\begin{equation}
d'_{m} = \frac{d_{1} + d_{2} + \cdots + d_{n}}{d_{m}}
.
\end{equation}
To reduce the total value of weights to 1 in (\ref{eqn:M_MLT}), $k_{m}$ is given as
\begin{equation}
\label{eqn:coefficient_k}
k_{m} = \frac{d'_{m}}{d'_{1}+d'_{2}+\cdots+d'_{n}}
.
\end{equation}
Note that, in (\ref{eqn:coefficient_k}), $k_{m}$ will be infinite if input pixel ${\bm{P}}_{\rm{XYZ}}$ is equal to ${\bm{T}}_{m}$ (i.e. $d_{m}=0$). 
In this case, let $k_{m}$ be a value of 1, and let the other weights be a value of zero.
\subsection{Properties of proposed n-color balancing}
The proposed n-color balancing has the following properties.
\begin{enumerate}[(a)]
\setlength{\itemsep}{-2pt}
\item ${\bf{M}}_{\rm{NCB}}={\bf{M}}_{m}$, and ${\bm{P}}_{\rm{NCB}}={\bm{G}}_{m}$, if ${\bm{P}}_{\rm{XYZ}}={\bm{T}}_{m}$.
Hence, all target colors are perfectly adjusted to their ground truth colors.
\item When choosing white as a target color under n = 1, n-color balancing is reduced to white-balancing.
\item ${\bf{M}}_{\rm{A}}$ can be designed in a similar way to white-balancing, so n-color balancing is applied to any color appearance model such as Bradford's model.
\item Even when ${\bm{P}}_{\rm{XYZ}}$ does not correspond to any target colors, lighting effects on ${\bm{P}}_{\rm{XYZ}}$ can be reduced.\label{item:non-target}
\end{enumerate}
\begin{figure}[tb]
\begin{center}
  \includegraphics[keepaspectratio, scale=0.35]{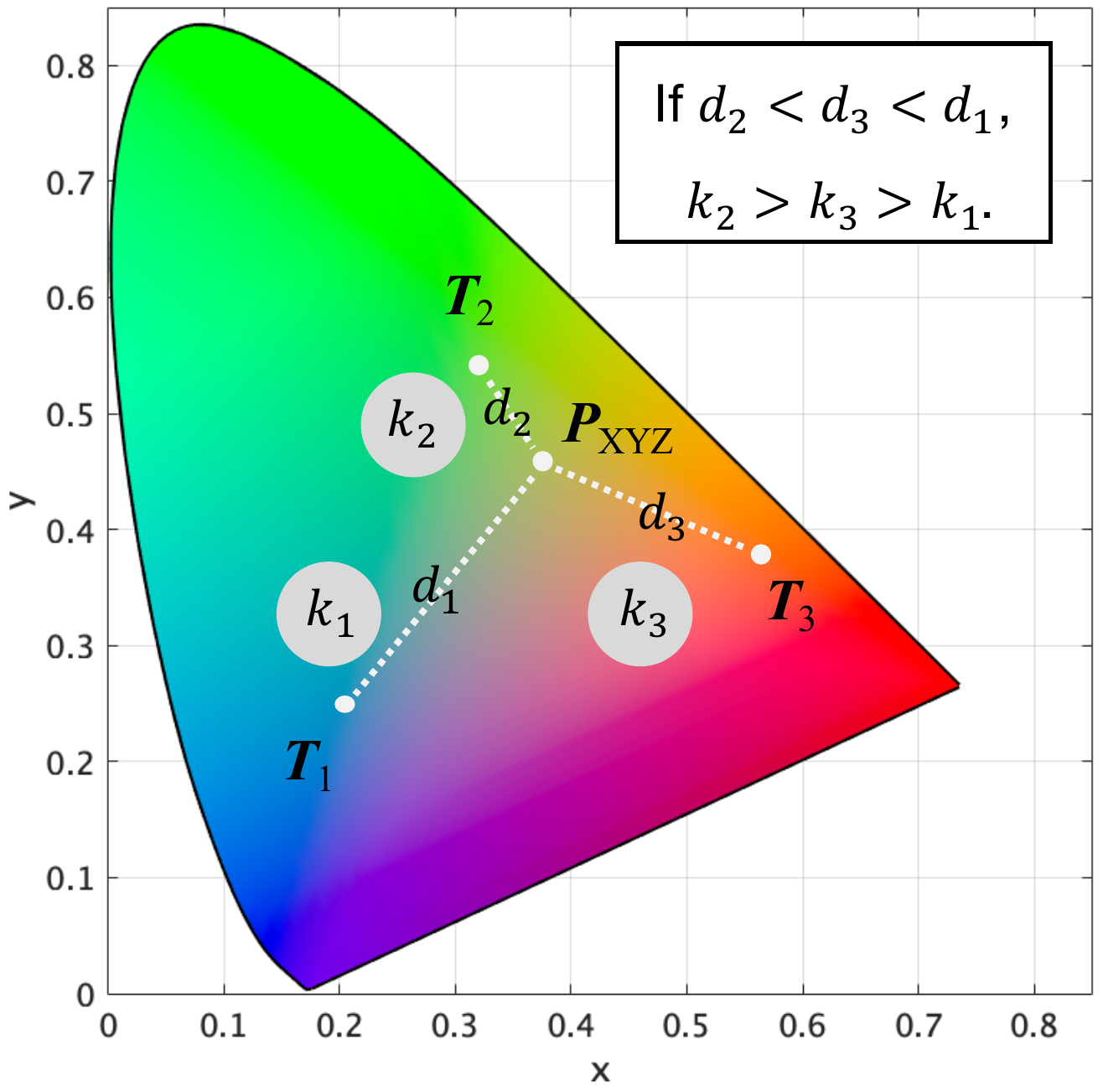}
\end{center}
\vspace{-0.5cm}
\caption{Example of distance $d_{m}$ between ${\bm{P}}_{\rm{XYZ}}$ and ${\bm{T}}_{m}$, and weight $k_{m}$.}
\label{fig:NCB_working_detail}
\end{figure}
%
As for (\ref{item:non-target}), if ${\bm{P}}_{\rm{XYZ}}$ has a closer color to ${\bm{T}}_{m}$, weight $k_{m}$ will be larger in accordance with  (\ref{eqn:XYZ_vector_distance}) (see Fig. \ref{fig:NCB_working_detail}).
Hence, n-color balancing can well reduce lighting effects on ${\bm{P}}_{\rm{XYZ}}$, even if ${\bm{P}}_{\rm{XYZ}}$ is not a target color.
\section{Experiment}
We conducted an experiment to confirm the effectiveness of the proposed method.
\subsection{Experimental setup}
We used the ColorChecker dataset prepared by Hemrit et al.\cite{ColorCheckerRECommended}, originally published by Gehler et al. in 2008\cite{GehlerShiDataset}. 
The dataset consists of 568 images capturing a color rendition chart, where the mean pixel value of each patch in the color chart is provided. 
The performance of each color balance adjustment was evaluated by using the reproduction angular error between an adjusted mean-pixel value and the ground truth one of each patch as in\cite{ReproductionError}. 
The score (or error) is calculated as
\begin{equation}
Err_{i} = {\rm{cos}}^{-1} \left( \frac{ {{\bm{P}}_{i}} \cdot {{\bm{Q}}_{i}} }{ \|{{\bm{P}}_{i}}\| \|{{\bm{Q}}_{i}}\| } \right),
\end{equation}
where $Err_{i}$ is the error of patch $i$, ${\bm{P}}_{i}$ is the mean pixel value of adjusted patch $i$, and ${{\bm{Q}}_{i}}$ is the mean pixel value of the ground truth. 
The ground truth image selected in this experiment was an image having the closest white to that of the CIE standard illuminant d65 (see Fig. \ref{fig:exp_results}).
\subsection{Experimental results}
\subsubsection*{A. Comparison with conventional methods}
%
\begin{figure}[htb]
\begin{center}
\includegraphics[keepaspectratio, scale=0.35]{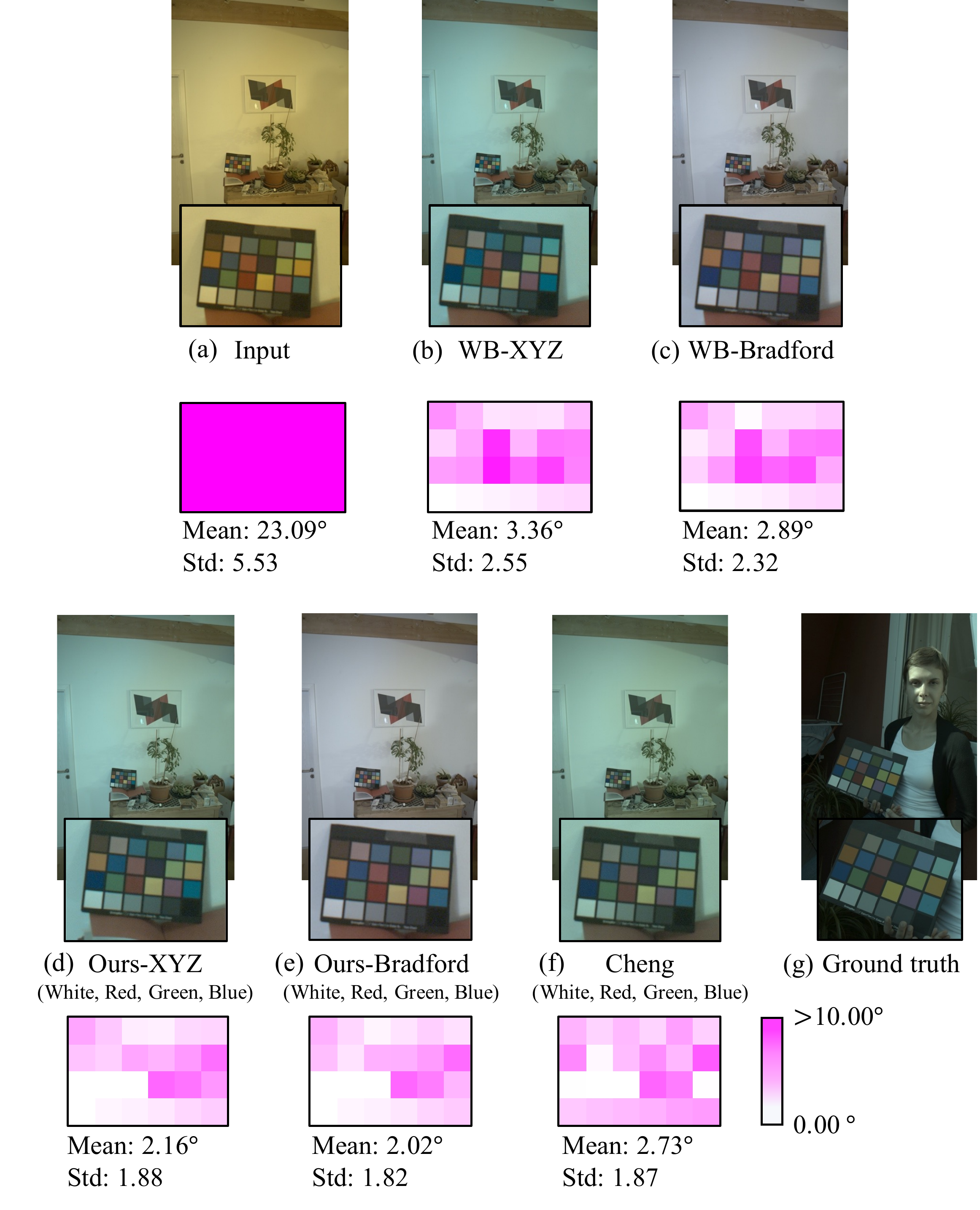}
\end{center}
\vspace{-0.5cm}
\caption{Example of adjusted images.
Difference in patch images between adjusted patches and ground truth ones, and mean error (Mean) and standard variation (Std) of 24 patches are shown below each image.}
\label{fig:exp_results}
\end{figure}
%
%
We used 551 images including a ground truth image in this experiment. 
Figure \ref{fig:exp_results} shows an example of images adjusted from an image in the dataset. 
In this experiment, two white balance adjustments, that is, white-balancing with XYZ scaling (WB-XYZ) and white-balancing with Bradford's model (WB-Bradford), our two methods, that is, n-color balancing with XYZ scaling (NCB-XYZ) and n-color balancing with Bradford's model (NCB-Bradford), and Cheng's method (Cheng) were applied to images including lighting effects, where white, red, green, and blue patches were chosen as target colors under n = 4 for Cheng's and the proposed methods.
As shown in Fig. \ref{fig:exp_results}, the proposed method with Bradford's model outperformed the other methods in terms of both the values of mean error and standard variation of 24 patches.
\subsubsection*{B. Evaluation of lighting effect on all patches}
%
%
\begin{table}[tb]
\caption{Mean error (Mean) and standard variation (Std) of $Err_{i}$ in every patch (deg). Patches 1 to 24 correspond to patches with different color. 
13, 14, 15, and 19 indicate blue, green, red, and white, respectively, for example, where (19) indicating white is a target color.}
\label{table:every_patch_results}
\begin{center}
\scalebox{0.51}{
\begin{tabular}{c|cc|cc|cc|cc|cc|cc}
\hline
\noalign{\vskip.5mm}
\multirow{3}{*}{Patch}  & \multicolumn{2}{c|}{Input} & \multicolumn{2}{c|}{WB-XYZ} & \multicolumn{2}{c|}{WB-Bradford} & \multicolumn{2}{c|}{NCB-XYZ} & \multicolumn{2}{c|}{NCB-Bradfrod} & \multicolumn{2}{c}{Cheng} \\ 
& \multicolumn{2}{c|}{} & \multicolumn{2}{c|}{(19)} & \multicolumn{2}{c|}{(19)} & \multicolumn{2}{c|}{(13, 14, 15, 19)} & \multicolumn{2}{c|}{(13, 14, 15, 19)} & \multicolumn{2}{c}{(13, 14, 15, 19)} \\ \cline{2-13}
& Mean & Std &Mean & Std & Mean & Std & Mean & Std & Mean & Std & Mean & Std  \\ \hline
1	&	8.602	&	6.026	&	1.939	&	1.472	&	1.755	&	1.385	&	0.907	&	0.948	&	0.831	&	0.879	&	\bf0.782	&	\bf0.733	\\
2	&	8.501	&	6.073	&	1.550	&	0.706	&	1.334	&	0.560	&	0.903	&	0.640	&	0.779	&	\bf0.537	&	\bf0.713	&	0.659	\\
3	&	7.802	&	6.821	&	0.903	&	0.719	&	\bf0.747	&	\bf0.524	&	1.488	&	0.900	&	1.191	&	0.727	&	1.356	&	0.907	\\
4	&	8.352	&	4.934	&	1.877	&	1.124	&	1.965	&	1.113	&	0.874	&	0.712	&	0.933	&	0.750	&	\bf0.647	&	\bf0.644	\\
5	&	8.017	&	7.222	&	\bf0.780	&	0.504	&	0.818	&	\bf0.400	&	1.270	&	0.787	&	1.136	&	0.607	&	1.511	&	1.110	\\
6	&	7.997	&	5.950	&	0.832	&	0.871	&	\bf0.700	&	0.740	&	0.952	&	0.853	&	0.751	&	0.740	&	1.054	&	\bf0.655	\\ \hline
7	&	6.167	&	3.240	&	1.977	&	1.070	&	1.826	&	1.035	&	\bf0.937	&	0.734	&	1.257	&	\bf0.650	&	3.602	&	1.555	\\
8	&	5.618	&	6.665	&	2.905	&	1.295	&	2.426	&	0.981	&	2.114	&	0.905	&	1.773	&	0.703	&	\bf0.720	&	\bf0.573	\\
9	&	9.997	&	6.585	&	3.951	&	2.116	&	3.495	&	1.808	&	1.611	&	0.981	&	1.442	&	\bf0.911	&	\bf1.261	&	0.977	\\
10	&	8.145	&	7.776	&	\bf1.394	&	0.874	&	1.448	&	\bf0.825	&	1.652	&	1.013	&	1.598	&	0.916	&	1.515	&	1.422	\\
11	&	6.245	&	3.359	&	1.417	&	1.077	&	1.565	&	1.026	&	\bf1.257	&	0.858	&	1.324	&	\bf0.794	&	2.319	&	1.250	\\
12	&	5.173	&	2.590	&	1.482	&	1.124	&	1.658	&	1.109	&	\bf1.426	&	1.164	&	1.739	&	\bf1.065	&	4.345	&	1.867	\\ \hline
13	&	5.172	&	6.130	&	3.175	&	1.989	&	2.561	&	1.482	&	\bf0.000	&	\bf0.000	&	\bf0.000	&	\bf0.000	&	0.011	&	0.032	\\
14	&	7.628	&	4.472	&	1.682	&	1.170	&	1.737	&	1.137	&	\bf0.000	&	\bf0.000	&	\bf0.000	&	\bf0.000	&	0.225	&	0.288	\\
15	&	9.862	&	5.521	&	5.085	&	2.796	&	4.564	&	2.574	&	\bf0.000	&	\bf0.000	&	\bf0.000	&	\bf0.000	&	0.004	&	0.014	\\
16	&	4.813	&	2.492	&	\bf1.354	&	1.309	&	1.545	&	1.231	&	1.924	&	1.283	&	2.048	&	\bf1.182	&	4.475	&	2.017	\\
17	&	10.679	&	8.003	&	3.208	&	1.727	&	3.061	&	1.572	&	\bf2.288	&	1.464	&	2.293	&	\bf1.362	&	2.984	&	1.468	\\
18	&	6.628	&	5.974	&	2.133	&	1.742	&	1.802	&	1.311	&	2.538	&	1.585	&	2.110	&	1.243	&	\bf0.952	&	\bf0.655	\\ \hline
19	&	8.011	&	6.409	&	\bf0.000	&	\bf0.000	&	\bf0.000	&	\bf0.000	&	\bf0.000	&	\bf0.000	&	\bf0.000	&	\bf0.000	&	0.777	&	0.750	\\
20	&	8.362	&	6.469	&	0.367	&	0.173	&	0.369	&	\bf0.172	&	\bf0.322	&	0.181	&	0.325	&	0.180	&	1.067	&	0.842	\\
21	&	8.479	&	6.504	&	0.522	&	0.278	&	0.524	&	0.274	&	\bf0.478	&	0.271	&	0.483	&	\bf0.268	&	1.189	&	0.885	\\
22	&	8.799	&	6.569	&	0.910	&	0.474	&	0.908	&	0.465	&	\bf0.847	&	0.410	&	0.850	&	\bf0.406	&	1.510	&	0.951	\\
23	&	8.800	&	6.813	&	0.953	&	0.905	&	0.950	&	0.896	&	0.850	&	\bf0.763	&	\bf0.847	&	0.765	&	1.555	&	1.241	\\
24	&	8.900	&	7.245	&	1.374	&	1.683	&	1.365	&	1.672	&	1.214	&	\bf1.492	&	\bf1.209	&	1.496	&	1.733	&	1.874	\\ \hline
Total&	\multirow{2}{*}{7.781}	&	\multirow{2}{*}{6.209}	&	\multirow{2}{*}{1.741}	&	\multirow{2}{*}{1.749}	&	\multirow{2}{*}{1.630}	&	\multirow{2}{*}{1.544}	&	\multirow{2}{*}{1.077}	&	\multirow{2}{*}{1.136}	&	\multirow{2}{*}{\bf1.038}	&	\multirow{2}{*}{\bf1.043}	&	\multirow{2}{*}{1.513}	&	\multirow{2}{*}{1.628}	\\
average&&&&&&&&&&&&	\\	
\noalign{\vskip.5mm}
\hline
\end{tabular}
}
\end{center}
\end{table}
%
In Table \ref{table:every_patch_results}, the proposed method is compared with the conventional methods in terms of mean error (Mean) and standard variation (Std) of $Err_{i}$ for every patch, where 13, 14, 15, and 19 indicate blue, green, red, and white, respectively. 
From Table \ref{table:every_patch_results}, the properties and performances of the proposed method are represented as follows.
\begin{itemize}
\setlength{\itemsep}{-2pt}
\item Our method perfectly adjusted the target colors (Mean = Std = 0).
\item NCB-XYZ outperformed WB-XYZ and NCB-Bradford outperformed WB-Bradford (see total average).
\item NCB-Bradford outperformed Cheng in terms of both Mean and Std.
\item Our methods were also effective in the correction of non-target colors.
\end{itemize}

In addition, the proposed method (n-color balancing) was already confirmed to be effective even under various target color selections. 
The performance had a similar trend to that with the selection of white, red, green, and blue as shown in Fig. \ref{fig:exp_results} and Table \ref{table:every_patch_results}.
\section{Conclusion}
We proposed a novel multi-color balance adjustment, called ``n-color balancing,'' for color constancy. 
n-color balancing is designed by using n matrices calculated from target colors and corresponding ground truth colors. 
It also allows us not only to perfectly adjust all target colors but also to correct colors other than the target colors. 
When choosing white as a target color under n = 1, n-color balancing is reduced to white-balancing. 
In addition, n-color balancing can be applied to any color appearance model such as Bradford's model. 
In an experiment, the proposed method was demonstrated to outperform the conventional white and multi-color balance adjustments.
%
%
\bibliographystyle{IEEEbib}
\bibliography{references}

\begin{thebibliography}{10}

\bibitem{Color_Constancy_effect_on_DNN_2019_ICCV}
M.~Afifi and M.~S. Brown,
\newblock ``What else can fool deep learning? addressing color constancy errors
  on deep neural network performance,''
\newblock in {\em Proceedings of the IEEE Conference on Computer Vision
  (ICCV)}, Oct. 2019, pp. 243--252.

\bibitem{Fast_Soft_Segmentation_2020_CVPR}
N.~Akimoto, H.~Zhu, Y.~Jin, and Y.~Aoki,
\newblock ``Fast soft color segmentation,''
\newblock in {\em Proceedings of the IEEE Conference on Computer Vision and
  Pattern Recognition (CVPR)}, Jun. 2020, pp. 8274--8283.

\bibitem{kinoshita-san-Hue}
Y.~{Kinoshita} and H.~{Kiya},
\newblock ``Hue-correction scheme based on constant-hue plane for
  deep-learning-based color-image enhancement,''
\newblock {\em IEEE Access}, vol. 8, pp. 9540--9550, Jan. 2020.

\bibitem{kinoshita-san-APSIPA-HUE}
Y.~Kinoshita and H.~Kiya,
\newblock ``Hue-correction scheme considering ciede2000 for color-image
  enhancement including deep-learning-based algorithms,''
\newblock {\em APSIPA Transactions on Signal and Information Processing}, vol.
  9, no. e19, pp. 1--10, Sep. 2020.

\bibitem{kinoshita-san-Semantic_2019}
Y.~{Kinoshita} and H.~{Kiya},
\newblock ``Scene segmentation-based luminance adjustment for multi-exposure
  image fusion,''
\newblock {\em IEEE Transactions on Image Processing}, vol. 28, no. 8, pp.
  4101--4116, Aug. 2019.

\bibitem{ChengsBeyondWhite}
D.~{Cheng}, B.~{Price}, S.~{Cohen}, and M.~S. {Brown},
\newblock ``Beyond white: Ground truth colors for color constancy correction,''
\newblock in {\em Proceedings of the IEEE Conference on Computer Vision
  (ICCV)}, Dec. 2015, pp. 298--306.

\bibitem{Max_RGB}
E.~H. Land and J.~J. McCann,
\newblock ``Lightness and retinex theory,''
\newblock {\em Journal of the Optical Society of America}, vol. 61, no. 1, pp.
  1--11, Jan. 1971.

\bibitem{Grey_World_Theory}
G.~Buchsbaum,
\newblock ``A spatial processor model for object colour perception,''
\newblock {\em Journal of the Franklin Institute}, vol. 310, no. 1, pp. 1--26,
  Jul. 1980.

\bibitem{Grey_Edge}
J.~{van de Weijer}, T.~{Gevers}, and A.~{Gijsenij},
\newblock ``Edge-based color constancy,''
\newblock {\em IEEE Transactions on Image Processing}, vol. 16, no. 9, pp.
  2207--2214, Aug. 2007.

\bibitem{Cheng_PCA}
D.~Cheng, D.~K. Prasad, and M.~S. Brown,
\newblock ``Illuminant estimation for color constancy: why spatial-domain
  methods work and the role of the color distribution,''
\newblock {\em Journal of the Optical Society of America}, vol. 31, no. 5, pp.
  1049--1058, May 2014.

\bibitem{APAP_bias_illumination_estimation}
M.~{Afifi}, A.~{Punnappurath}, G.~D. {Finlayson}, and M.~S. {Brown},
\newblock ``As-projective-as-possible bias correction for illumination
  estimation algorithms,''
\newblock {\em Journal of the Optical Society of America}, vol. 36, no. 1, pp.
  71--78, Jan. 2019.

\bibitem{Deep_WB_Editing}
M.~Afifi and M.~S. Brown,
\newblock ``Deep white-balance editing,''
\newblock in {\em Proceedings of the IEEE Conference on Computer Vision and
  Pattern Recognition (CVPR)}, Jun. 2020, pp. 1397--1406.

\bibitem{vonKriesOriginal}
J.~von Kries,
\newblock ``Beitrag zur physiologie der gesichtsempfindung,''
\newblock {\em Archives of Anatomy and Physiology}, pp. 503--524, 1878.

\bibitem{BradfordOriginal}
K.~M. {Lam},
\newblock {\em Metamerism and colour constancy},
\newblock Ph.D. thesis, University of Bradford, 1985.

\bibitem{Maloney:color_constacy}
L.~T. Maloney and B.~A. Wandell,
\newblock ``Color constancy: a method for recovering surface spectral
  reflectance,''
\newblock {\em Journal of the Optical Society of America}, vol. 3, no. 1, pp.
  29--33, Jan. 1986.

\bibitem{Non_diagonal_Color_Correction}
B.~{Funt} and H.~{Jiang},
\newblock ``Nondiagonal color correction,''
\newblock in {\em Proceedings of the IEEE Conference on Image Processing
  (ICIP)}, Sep. 2003, pp. 481--484.

\bibitem{Non_diagonal_Color_Correction_PCA_Analysis}
C.~Huang and D.~Huang,
\newblock ``{A study of non-diagonal models for image white balance},''
\newblock in {\em Image Processing: Algorithms and Systems XI}, Feb. 2013, vol.
  8655, pp. 384--395.

\bibitem{CAM16original}
C.~Li, Z.~Li, Z.~Wang, Y.~Xu, M.~R. Luo, G.~Cui, M.~Melgosa, M.~H. Brill, and
  M.~Pointer,
\newblock ``Comprehensive color solutions: Cam16, cat16, and cam16-ucs,''
\newblock {\em Color Research \& Application}, vol. 42, no. 6, pp. 703--718,
  Jun. 2017.

\bibitem{Karaimer_2018_CVPR}
H.~C. Karaimer and M.~S. Brown,
\newblock ``Improving color reproduction accuracy on cameras,''
\newblock in {\em Proceedings of the IEEE Conference on Computer Vision and
  Pattern Recognition (CVPR)}, Jun. 2018, pp. 6440--6449.

\bibitem{Akazawa_3CB}
T.~Akazawa, Y.~Kinoshita, and H.~Kiya,
\newblock ``Multi-color balancing for correctly adjusting the intensity of
  target colors,''
\newblock in {\em Proceedings of 2021 IEEE 3rd Global Conference on Life
  Sciences and Technologies (LifeTech)}, 2021, pp. 8--12.

\bibitem{ColorCheckerRECommended}
G.~Hemrit, G.~D. Finlayson, A.~Gijsenij, P.~Gehler, S.~Bianco, B.~Funt,
  M.~Drew, and L.~Shi,
\newblock ``Rehabilitating the colorchecker dataset for illuminant
  estimation,''
\newblock {\em Color and Imaging Conference}, vol. 2018, no. 1, pp. 350--353,
  Nov. 2018.

\bibitem{Lambertian_model}
J.H. {Lambert},
\newblock {\em Photometria, sive de Mensura et Gradibus Luminis, Colorum et
  Umbrae},
\newblock Klett, 1760.

\bibitem{brucelindbloom}
B.~{Lindbloom}.,
\newblock ``Chromatic adaptation.,''
  \url{http://www.brucelindbloom.com/index.html?Eqn\_ChromAdapt.html},
\newblock accessed Dec. 11, 2020.

\bibitem{GehlerShiDataset}
P.~V. {Gehler}, C.~{Rother}, A.~{Blake}, T.~{Minka}, and T.~{Sharp},
\newblock ``Bayesian color constancy revisited,''
\newblock in {\em Proceedings of the IEEE Conference on Computer Vision and
  Pattern Recognition (CVPR)}, Aug. 2008.

\bibitem{ReproductionError}
G.~D. Finlayson and R.~Zakizadeh,
\newblock ``Reproduction angular error: An improved performance metric for
  illuminant estimation,''
\newblock in {\em Proceedings of the British Machine Vision Conference}, Sep.
  2014.

\end{thebibliography}

\end{document}